%% file: main.tex
\documentclass[lettersize,journal]{IEEEtran}
\usepackage[utf8]{inputenc}
\usepackage[T1]{fontenc}
\usepackage{hyperref}

\usepackage{graphicx}
\usepackage[capitalise]{cleveref}
\usepackage{multirow}
\usepackage{array}
\usepackage{siunitx}

\begin{document}
%
\title{MeciFace: Mechanomyography and Inertial Fusion-based Glasses for Edge Real-Time Recognition of Facial and Eating Activities}
%
%
%

\author{Hymalai~Bello,
        Sungho~Suh,~\IEEEmembership{Member,~IEEE,},
        Bo~Zhou,~\IEEEmembership{Member,~IEEE,},
        Paul~Lukowicz,~\IEEEmembership{Member,~IEEE,}
\IEEEcompsocitemizethanks{\IEEEcompsocthanksitem Hymalai Bello, Sungho Suh, Bo Zhou, and Paul Lukowicz are with the Department of Embedded Intelligence, German Research Center for Artificial Intelligence and the University of Kaiserslautern-Landau, 67663, Kaiserslautern, Germany.\protect 
\IEEEcompsocthanksitem Corresponding Author: Hymalai Bello (Hymalai.Bello@dfki.de)
}     
}

%
%

    \markboth{IEEE Transactions on Consumer Electronics}%
{Shell \MakeLowercase{\textit{et al.}}: Bare Demo of IEEEtran.cls for IEEE Journals}
%



\maketitle

\begin{abstract}
The increasing prevalence of stress-related eating behaviors and their impact on overall health highlights the importance of effective and ubiquitous monitoring systems. 
In this paper, we present MeciFace, an innovative wearable technology designed to monitor facial expressions and eating activities in real-time on-the-edge (RTE).
MeciFace aims to provide a low-power, privacy-conscious, and highly accurate tool for promoting healthy eating behaviors and stress management. 
We employ lightweight convolutional neural networks as backbone models for facial expression and eating monitoring scenarios. 
The MeciFace system ensures efficient data processing with a tiny memory footprint, ranging from $11KB$ to $19KB$. 
During RTE evaluation, the system achieves an F1-score of $\geq86\%$ for facial expression recognition and 94\% for eating/drinking monitoring, for the RTE of unseen users (user-independent case).
\end{abstract}


%
\IEEEpeerreviewmaketitle

\input{Chapters/Introduction.tex}

\input{Chapters/RelatedWork.tex}	

\input{Chapters/Methods.tex}

\input{Chapters/ResultsDiscussion.tex}

\input{Chapters/Conclusion.tex}

\section*{Acknowledgment}
The research reported in this paper was partially supported by the German Federal Ministry of Education and Research (BMBF) in the project SocialWear (01IW20002) and Eghi (16SV8527).

\bibliographystyle{IEEEtran}

\bibliography{References}

%


\begin{IEEEbiography}[{\includegraphics[width=1in, height=1.25in, clip,keepaspectratio]{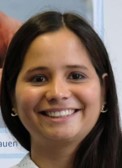}}]{Hymalai Bello}
		She received a master's degree in electrical engineering and computer science from the University of Kaiserslautern in 2017. She is currently working as a researcher in the Department of Embedded Intelligence at DFKI with Prof. Paul Lukowicz. Her main research interests are hardware and software co-design of multimodal sensor-based systems for human activity recognition. 
\end{IEEEbiography}
  \vspace{-10pt}
\begin{IEEEbiography}[{\includegraphics[width=1in, height=1.25in, clip,keepaspectratio]{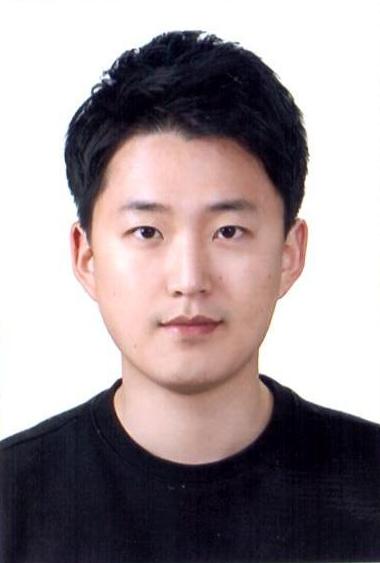}}]{Sungho Suh}
		is a Senior Researcher at the German Research Center for Artificial Intelligence (DFKI) in Germany since 2021. He received the Ph.D. degree in Computer Science at the Technische Universit{\"a}t Kaiserslautern, Germany in 2021, and the B.S. and M.S. degrees from the School of Electrical Engineering and Computer Science, Seoul National University, Seoul, South Korea, in 2009 and 2011, respectively. Before joining DFKI, he worked at KIST Europe in Germany for three years, and at Samsung Electro-Mechanics, Korea from 2011 to 2018. His research interests are machine learning algorithms, such as sensor data processing, computer vision, multimodal processing, and generative models, with a focus on industrial applications.
\end{IEEEbiography}
   \vspace{-10pt}
\begin{IEEEbiography}[{\includegraphics[width=1in, height=1.25in, trim={1cm 0cm 1cm 0mm},clip,keepaspectratio]{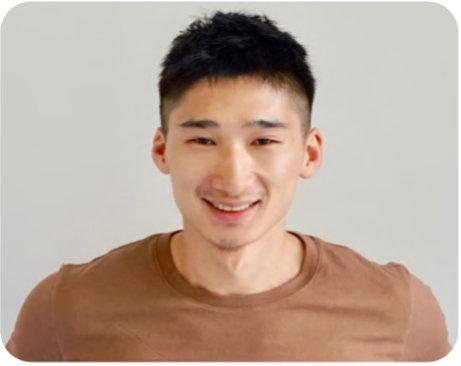}}]{Bo Zhou}
    is a Senior Researcher at the German Research Center for Artificial Intelligence (DFKI GmbH). His main research interests are sensor-based perception through hardware-software co-designed systems and sensor signal analysis. He received his Ph.D. in Computer Science at RPTU Kaiserslautern-Landau, and his M.S. degrees in Electronics from the University of Southampton, RPTU Kaiserslautern-Landau and NTNU.
\end{IEEEbiography}
       \vspace{-10pt}
\begin{IEEEbiography}[{\includegraphics[width=1in, height=1.25in, trim={5cm 0cm 5cm 0mm},clip, keepaspectratio]{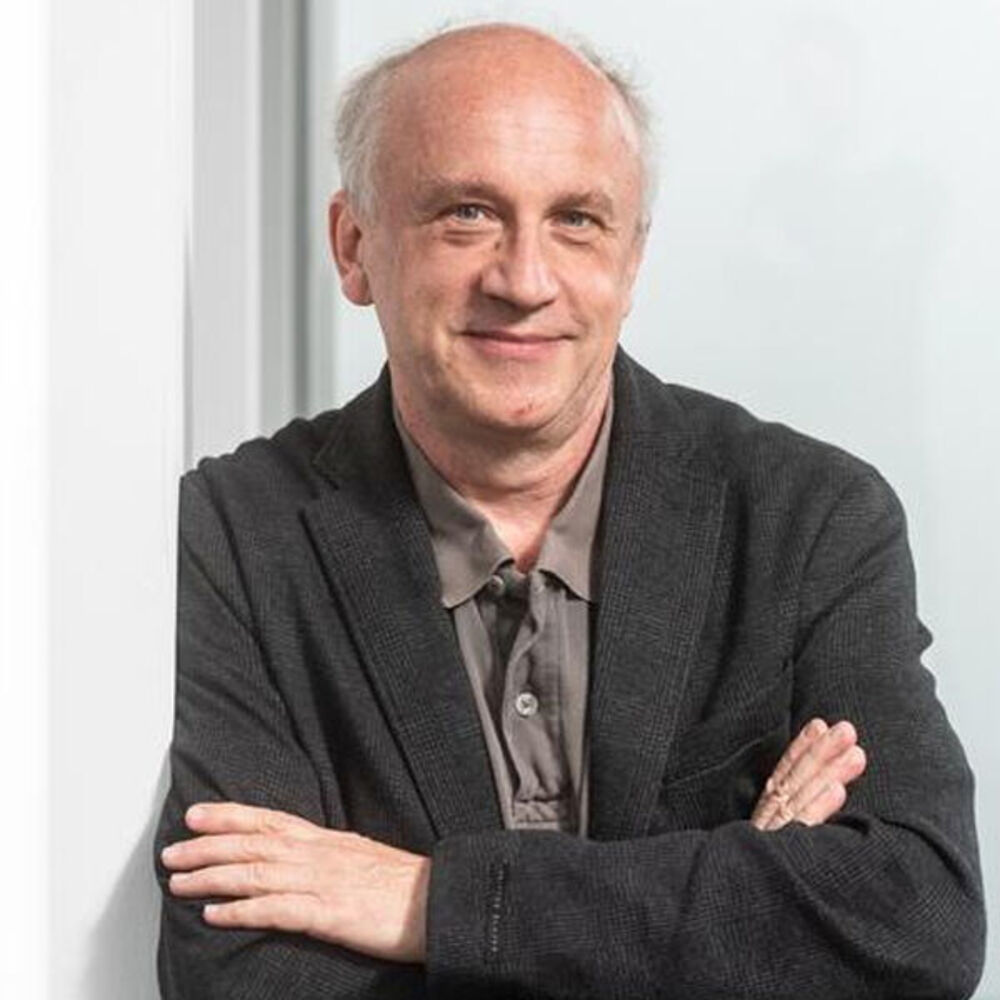}}]{Paul Lukowicz}
		is both Scientific Director at the German Research Center for Artificial Intelligence (DFKI GmbH) and Professor of Computer Science at RPTU Kaiserslautern-Landau in Germany since 2012 where he heads the Embedded Intelligence group. His research focuses on context-aware ubiquitous and wearable systems including sensing, pattern recognition, system architectures, models of large-scale self-organized systems, and applications in areas ranging from healthcare through Industry 4.0 to smart cities.
\end{IEEEbiography}

\end{document}

%% file: Chapters/Introduction.tex
\section{Introduction}
\label{sec:Intro}
Facial expression recognition and eating monitoring technologies have become increasingly essential in understanding stress-related eating behaviors and their impact on overall health \cite{foodStress}. 
Wearable devices offer a convenient and non-intrusive solution to detect potential health issues, such as binge eating, stress-related overeating, and anorexia \cite{stresseating}, and monitor these behaviors, catering to individuals with specific dietary restrictions, such as diabetic patients or those with food sensitivities. 
In addition, wearables can help individuals develop coping mechanisms to manage stress and maintain a healthy lifestyle.

The use of glasses for human activity recognition (HAR) in wearable technologies is widespread due to their ubiquity and strategic position in front of the user's face. 
In general, glasses-based wearables offer a comprehensive approach to capturing visual cues, tracking eye movements, integrating sensors, providing real-time information, and enabling hands-free use. 
These aspects contribute significantly to accurate and efficient activity recognition \cite{FaceCommands}, \cite{facerecglasses, EmotionCamEDAPPG, PhotoReflectiveGlass, KissglassIMUOculo, Capglasses, MicArrayActiveFacialGlasses}. 
Commercial solutions, such as OCOsense (Emteq Labs), have already been introduced, utilizing optical-flow, inertial, pressure, and microphone sensors to monitor facial expressions and emotions \cite{OCOsense}.

However, existing state-of-the-art glasses-based wearables typically rely on external devices, such as computers, servers, or smartphones/tablets for real-time data processing and inference. 
The distribution of power consumption, latency, and memory can limit the efficiency of the system.
Handling data across multiple devices can become a privacy and security concern. 
To address these limitations, in this paper, we introduce MeciFace, a real-time solution that performs facial activity recognition and eating/drinking gesture detection on-the-edge. 
By embedding data acquisition, signal processing, and inference within the MeciFace hardware, the system minimizes reliance on external devices.

\begin{figure*}
    \centering
    \includegraphics[width=\textwidth]{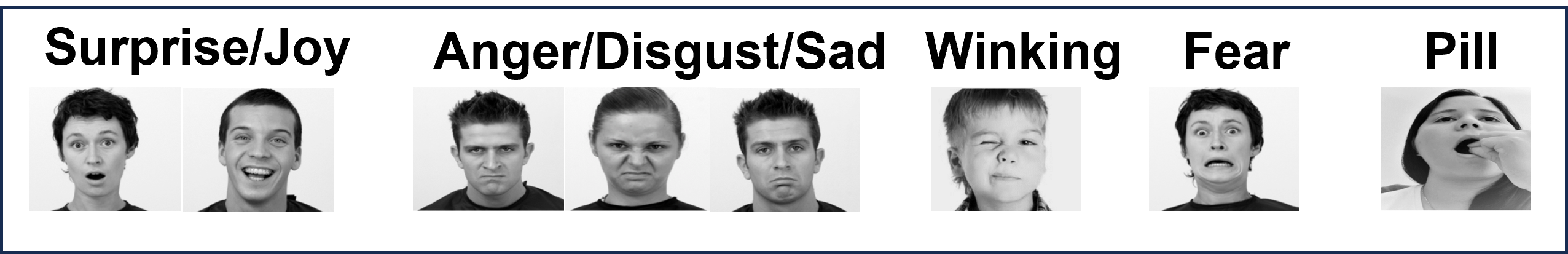}
    \caption{Facial Muscle Activities Dictionary; 6 Facial Expressions from Warsaw Set of Emotional Facial Expression Photoset \cite{Warsaw} and 2 Gestures from \cite{bello2023inmyface}. Taking a Pill Facial Muscle Movement is Included to Differentiate Eating/Drinking Episode with the Sporadic Gesture of Touching Face/Mouth.}  
    \label{fig:FacialDic}
\end{figure*}

The MeciFace system utilizes neural network models (NN) deployed on a microcontroller (MCU) using the TensorFlow Lite for microcontrollers framework. 
The proposed system fuses information from inertial and mechanomyography (MMG) sensors, ensuring privacy and low power consumption while achieving robust recognition performance \cite{bello2023inmyface}.
The facial expression dictionary is defined in \cref{fig:FacialDic}, in addition to the null/else class as static face \cite{bello2023inmyface, bello2020facial}. 
The gesture dictionary is based on the Warsaw Set of Emotional Facial Expression Pictures (WSEFEP). 
The Warsaw dataset provides a validated baseline of photographs of genuine facial expressions. 
These pictures serve as a guide for participants to imitate the facial gesture as closely as possible while wearing the MeciFace. 
In particular, the gesture of taking a pill is included to differentiate eating/drinking episodes from the sporadic gesture of touching the face/mouth. 
The eating scenario-related classes are eating, drinking, and null to extend the work in \cite{zhou2020snacap}.  

The main contributions of our approach can be summarized as follows:
\begin{itemize}
    \item We present MeciFace, a state-of-the-art real-time solution for facial activity related to facial expressions and eating/drinking gestures that uses a fusion of mechanomyography and inertial sensing, providing flexibility, low power consumption, and cost-effectiveness, with potential future applications such as monitoring sporadic episodes of emotional eating.  
    \item We employ lightweight neural network models to ensure a low memory footprint, providing an embedded and sustainable solution.  
    \item We propose a hierarchical multimodal fusion to reduce energy consumption and increase robustness against the null class, in which the first stage detects motions and recognizes a non-null facial gesture using an MMG model. Then, using an inertial model, the second stage recognizes the dictionary in \cref{fig:FacialDic}.
    \item The hierarchical multimodal fusion is extended for the case of eating/drinking monitoring. The first stage discriminates between null and eating/drinking categories with an MMG model. The second stage employs an inertial model to classify between eating and drinking.  
    \item Our work is a first step towards a ubiquitous system that monitors facial expressions and eating/drinking episodes to add contextual information from both scenarios, relevant to detecting stress-triggered eating episodes.  
\end{itemize}

 The paper is organized as follows; \cref{sec:RelatedWork} presents a summary of the state of the art for wearable to recognize facial and eating activities. \cref{sec:methods} provides detailed information on the hardware prototype, and introduces the details of the proposed multimodal fusion for facial activity recognition, and \cref{sec:Results} presents the experimental results and discussion. Finally, \cref{sec:Conclusion} concludes the paper.

%% file: Chapters/RelatedWork.tex
\section{Related Work}
\label{sec:RelatedWork}
Head-mounted accessories are the default option to monitor facial activity or eating scenarios. 
Wearable accessories such as glasses, earphones, headphones, or caps have proved to be reliable ubiquitous solutions for human activity recognition by many researchers \cite{bello2020facial, zhou2020snacap, bello2023inmyface, Earcommand, Ihearken}.

Currently, headphones or headsets are widely used as accessories for HAR. 
In \cite{ExpressEar}, the authors reused a commercial headset with IMU to monitor facial action units (FAUs).
In \cite{Facelistener}, an earpiece has been transformed into an acoustic detection device. 
The device sends ultrasound signals around the facial area.
The idea is to capture changes in the ultrasonic wave as a result of skin deformations caused by facial muscle movements with different facial expressions.
Ear-mounted cameras for facial contour reconstruction are proposed in \cite{C-face}. 
Ear-mounted devices offer a convenient option for facial and eating monitoring due to their portability. 
The headsets are mostly used in specific situations, such as commuting to work, physical exercise, or leisure activities. 
On the other hand, eyeglasses have become a fashion statement, with various frame styles to suit different personal preferences and styles, and are often worn throughout the day. 
Although our sensing modalities (inertial and mechanomyography) can also be deployed on ear-mounted devices, we favor the use of glasses due to user acceptance and a direct field of view of the contextual activities performed by the user.

In this section, we focus on a summary of related work that includes wearable glasses-based designs to recognize facial and eating activities. 
Many researchers in the wearable community proposed glasses-based design solutions for HAR. 
The authors in \cite{UHAR} combined a commercial eye-tracker with an IMU in a glasses-based design to derive contextual information by performing seven physical or cognitive tasks (talking, solving, reading, watching a video, typing, walking, and cycling). 
OCOSense \cite{OCOsense} presented a commercial solution to monitor facial activity and emotion based on optomyography, IMU, pressure, and microphone sensors. 
In \cite{Fitbyte}, inertial and optical sensors are combined to reliably detect food intake events, and with a camera, the system opportunistically captures pictures of the food as the user consumes it.
The system achieved an F1 score of 89\% in detecting eating and drinking in noisy environments. 
In \cite{ChewingGlasses}, the authors employed a glasses-based design for chewing monitoring using piezoelectric and inertial information with results up to 91 \%. 
Thus, the fusion of mechanomyography and inertial sensors for eating-related activities is a feasible option to explore further, for example, eating/drinking and the null class. 
The null class can include walking, talking, standing/sitting down, picking cutlery, and working on the PC, among others.

\begin{table}[t]
\caption{Comparison with State-of-the-art Glasses-based Real-Time Solution for Facial Activity Recognition} 
\centering
\begin{tabular}{ll}
\hline
Research & Highlights\\
\hline 
\multirow{4}{*}{OCOSense \cite{OCOsense}} & \textbullet~~Optomyographic, IMU and Altimeter    \\
                             & \textbullet~~Facial and Head Movements    \\
                             & \textbullet~~Real-Time Smartphone or Tablet   \\
                             & \textbullet~~Commercially Available\\
\hline
\multirow{3}{*}{SPIDERS+ \cite{spiders+}} & \textbullet~~IR Camera, Proximity Sensor and IMU \\
                                & \textbullet~~Facial Expressions\\
                                & \textbullet~~Semi-Real-Time PC, Cloud and Smartphone\\
\hline
\multirow{3}{*}{Kwon et al. \cite{Kwonemotion}}  
                            & \textbullet~~Images \\
                            &\textbullet~~Electrodermal Activity(EDA)\\ 
                            &\textbullet~~Photoplethysmogram(PPG)  \\
                            & \textbullet~~Emotional State   \\
                            & \textbullet~~Real-Time PC \\          
\hline
\multirow{3}{*}{EchoSpeech \cite{echospeech}} & \textbullet~~Active Acoustic Sensing  \\
                                        & \textbullet~~Facial Skin Deformation  \\
                                        & \textbullet~~Real-Time Smartphone  \\
\hline
\multirow{3}{*}{MyDJ \cite{ChewingGlasses}} & \textbullet~~IMU and Piezoelectric  \\
                                        & \textbullet~~Chewing Recognition  \\
                                        & \textbullet~~Real-Time ($\geq$ 3 seconds) MCU  \\  
\hline
\end{tabular}
\label{table:Comparison}
\end{table}

Most of the state-of-the-art approaches reported offline recognition results. 
In the computer vision society, the authors in \cite{emotionnet} focused on the efficient design of NNs for real-time recognition of facial expressions. 
The author's solution is for images, and the number of parameters is around 232000.
A few cases involve real-time recognition. 
And, most of them rely on cloud-based or mobile phone-based online recognition, see \cref{table:Comparison}. 
For an on-the-edge solution, the primary restrictions include low memory (Flash and RAM), latency awareness, and low power consumption. 
The wearable condition adds requirements such as; reduced size and weight and a limited number of connections to guarantee comfort. 
This work presents a real-time on-the-edge solution using TensorFlow Lite for the MCU framework to deploy the NN models. 

%% file: Chapters/Methods.tex
\section{Methods}
\label{sec:methods}
\begin{figure*}
    \centering
    \includegraphics[width=\textwidth]{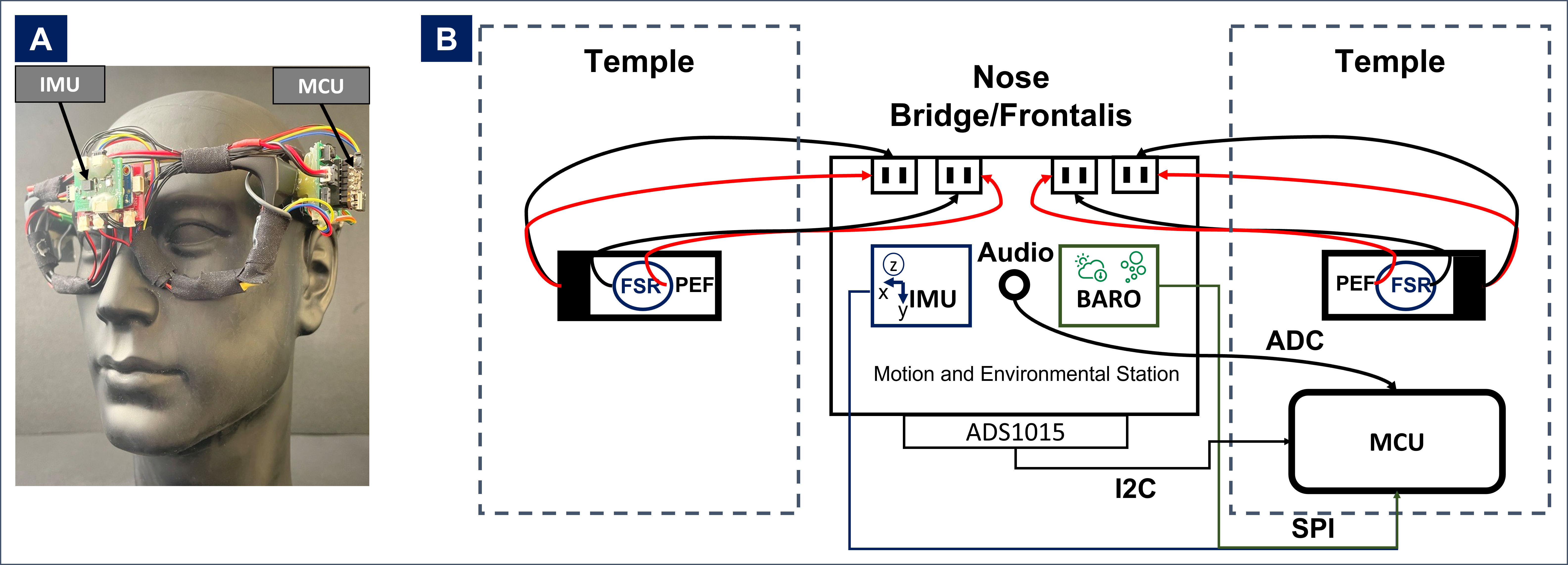}
    \caption{MeciFace Prototype \textbf{(A)}.
    Hardware Connections Blocks: Motion and Environmental Station on The Glasses' Nose Bridge with BNO085 (IMU),  SPH8878LR5H (Microphone) and BME688 (Barometer). On The Temples are The Force Sensitive Resistor (FSR), Piezoelectric Film (PEF), and QtPy ESP32 (MCU) \textbf{(B)}.}   
    \label{fig:HW}  
 \end{figure*}

The MeciFace prototype is shown in \cref{fig:HW}\textbf{A}. 
The microcontroller is a QTPy ESP32 from Adarfruit. 
The MCU is an ESP32-S3-Dual-Core 240MHz Tensilica with 8MB flash, 512KB SRAM, and Bluetooth low energy (BLE).
The prototype includes an SD Card, which is used as a data logger.  
The sensors are an inertial measurement unit (IMU-BNO085), an atmospheric pressure, environmental and gas sensor (BME688), an analog microphone (SPH8878LR5H), a force-sensitive resistor (FSR), and a piezo-electric film (PEF) for MMG, see \cref{table:Sensors}.

\begin{table*}[t]
\caption{Sensors Characteristics} 
\centering
\begin{tabular}{cllm{5em}l}
\hline
Sensor & Manufacturer & Dimensions(cm) & Weight (grams) & Benefits\\
\hline 
FSR
& Alpha MF01A-N-221-A01 
& 1.25 diameter 
& 0.26 
& Ultra-thin/flexible \\
PEF
& TE SDT1-028K shielded
&  4.45 x 1.97 x 0.32 
& 0.30 
& Low noise/shielded/flexible\\
IMU
& Bosch BNO085
& 0.38 x 0.52 x 0.11 
& 0.15
& Fused data, Auto calibration\\
Barometer
& Bosch BME688
& 3.0 x 3.0 x 0.9 mm³   
& 0.3
& Pressure and Gas Sensor with AI \\
Microphone
& Knowles SPH8878LR5H
& 0.35 x 0.27 x 0.13   
& 0.25
& Low Noise and Omnidirectional \\
\hline
\vspace{-20pt}
\end{tabular}
\label{table:Sensors}

\end{table*}

In \cref{fig:HW}\textbf{B}, the block connections diagram is presented. 
The IMU sensor connects to the MCU via a serial peripheral interface (SPI) bus. 
The hardware includes a pressure and environmental sensor (BME688) and analog audio (SPH8878LR5H). 
The data from the BME688 and SPH8878LR5H is not used in this work. 
Still, the option of monitoring environmental and audio data makes our design extendable for future analysis. 
FSR and PEF information is transferred to the MCU using an inter-integrated circuit (I2C) with the intermediate assistance of ADS1015. 
The ADS1015 is an analog to I2C converter. 
Converting analog signals to digital makes the system robust to subtle movements/motion artifacts, and reduces the signal's sensitivity to the distance between the sensor position and the MCU. 
Besides, it is also easier to add slaves to an I2C bus compared to adding more analog channels to the system. 
The sampling rate of the sensors is around 50 Hz.
The battery is a LiPo; 3.7V, 500mAh(3.5x3 cm). 

The IMU is on the nose bridge of the glasses to mimic the position of the temporal muscle in \cite{bello2023inmyface}. 
The IMU position is suitable to capture head displacement, cheek movements, and symmetrically sense vibrations on the glasses frame. 
The FSR and PEF are on the temples (right/left) muscles also used in \cite{bello2023inmyface}
The temple position is relevant to monitor masseter muscle-related movements. 
Masseter's movements include chewing, swallowing, and tongue sweeping for the case of eating activities and smiling or getting angry for the facial expression scenario.

\subsection{Multimodal Sensor Fusion}
As shown in \cref{fig:RT}\textbf{A} and in \cref{fig:RT}\textbf{B}, two collaborative models were deployed for the facial and eating monitoring applications. 
The first neural network model (NN) is the FSR-Piezo (MMG-model) with four channels as input. 
Two FSR and two Piezo channels complete the four inputs of the MMG model.
This model is used to distinguish the null class from activity detection.
The null class includes activities such as; walking, talking, standing/sitting down, picking cutlery, and working on the PC, among others. 
The output of the MMG model served as a trigger for the second model, the inertial model.
In the event of an activity being classified as non-null, the inertial model is activated.
The second model fused inertial information, including acceleration and orientation, as seven input channels.
Specifically, the input channels are linear accelerations (x, y, and z axes) and quaternions as orientation. 
For the case of facial expressions, the second model outputs are the classes in \cref{fig:FacialDic}.  
For the case of eating monitoring, the inertial model returns to eating/drinking classes. 
The hierarchical approach reduces the complexity of the models, leveraging the information fusion with lightweight NNs (11-19KB) to be deployed in tiny MCUs.
Using the MMG-model as a trigger signal, the power consumption remains below or equal to 0.55 Watts.

\subsection{Model}
The NN structure consists of a convolution (filters=3, kernel= 10, ReLu), a layer normalization, and batch normalization layers with max-pooling ((5,1)) and dropout (0.5), followed by a flattening layer, a fully connected (FC) layer of 10 and an FC with softmax. 
The NN optimizer is AdaDelta, with a learning rate of 0.9 and categorical cross-entropy (label smoothing 30\%) as a loss function. 
The metric to monitor during training is a recall at a precision of 0.9. 
This NN structure is used in the MMG and inertial models for both applications (Expressions/Eating). 
The training ran for 200 epochs with early stopping (patience 30 and restoring weights). 
The number of parameters of our NNs is $\leq3890$; thus it is a lightweight design and less susceptible to overfitting.
The NN models were trained using the TensorFlow/Keras 2.12.0 framework. 

\subsection{Ethical Agreement}
All participants signed an informed consent following the Declaration of Helsinki.
The ethical committee of Kaiserslautern University and the German Research Center for Artificial Intelligence have approved the study.
Participation was entirely voluntary and could be withdrawn at any time.
The participants did not receive any compensation for their participation. 
The subjects could deny answering questions if they feel uncomfortable in any way.
There are no risks associated with this user study. 
Discomforts or inconveniences will be minor and are not likely to happen. 
All data provided in this user study will be treated confidentially, will be saved encrypted, and cannot be viewed by anyone outside this research project unless separate permission is signed to allow it. 
The data in this study will be subject to the General Data Protection Regulation (GDPR) of the European Union (EU) and treated in compliance with the GPDR. 

\subsection{Evaluation Eating/Drinking Scenario}
Two groups of volunteers were recruited for the evaluation, for a total of ten participants. 
The training group and the test group have the same size and sex composition, but the participants do not overlap.  
For the training group, five volunteers (three female and two male) participated in the eating monitoring experiment.
The volunteers come from Germany, the Republic of Korea, China, and the United Kingdom, and range in age from 24 to 64 years old (mean 47).
The participants consumed their lunch or dinner without any restriction in a natural setting during four separate sessions.
It is important to note that participants were not forced to perform special activities or follow a script to ingest their food. 
Thus, the null activities are acquired in a natural/authentic setting. 
The four sessions per participant were recorded on different days, ensuring that our device was worn repeatedly.
For the eating/drinking case, the offline evaluation scheme was 4-fold cross-validation with a leaving-one-session-out. 
In addition, another group of five participants (testing group) was recruited for the real-time and on-the-edge evaluation. 
Therefore, the real-time evaluation was performed with another five participants (three female and two male), whose data were not used during model training.  
For the RTE assessment, volunteers come from India, Poland, USA, Germany, and Venezuela, and range in age from 25 to 34 years (mean 28,6).  
With this methodology, our results are user-independent and sex-balanced, with high cultural variability.

\subsection{Evaluation Facial Scenario}
For the facial scenario, one person mimicked (randomly-10 sessions) the dictionary in \cref{fig:FacialDic} while wearing the MeciFace.
A 10-fold cross-validation with a leave session out scheme was used. 
The ten sessions were on different days. 
Each session has four random tries per expression. 
The facial experiment is an extension of the previous work in \cite{bello2023inmyface}. 
In \cite{bello2023inmyface}, we fused MMG and inertial data to monitor facial expressions with a sports cap design and thirteen participants (offline evaluation). 
In this work, we focus on the real-time glasses-based idea implementation for a more ubiquitous/embedded solution.

\subsection{Real-time on-the-Edge Recognition (RTE)} 
\begin{figure*}
    \centering
    \includegraphics[width=\textwidth]{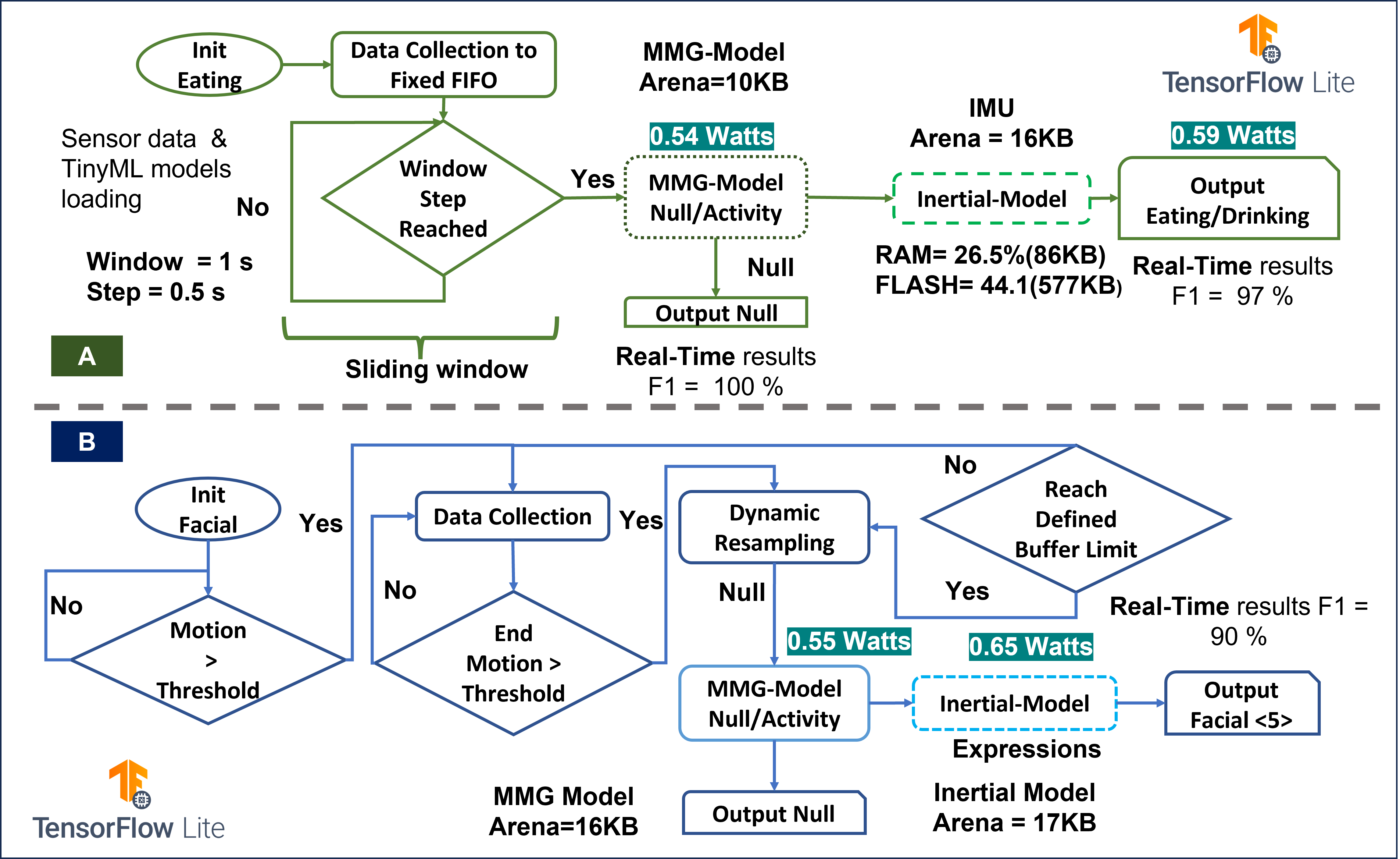}
    \caption{Real-Time and on-the-Edge Flow Diagram Implementation for the Eating/Drinking Scenario with the Two Stages Hierarchical Modeling; First Stage is the Mechanomyography-based Model (MMG-Model) to Detect Null/Activity. The Second Stage is the Inertial-Model to Classify Eating and Drinking Episodes by Window Size of One Second and Window Step of Half a Second \textbf{(A)}.
    Real-Time and on-the-Edge Flow Diagram Implementation for the Facial Expressions Scenario with Motion Threshold Detection and Two Stages Hierarchical Modeling; The First Stage is the MMG-Model to detect Null/Activity. The Second Stage is the Inertial-based Model to Classify the Facial Movements Dictionary in \cref{fig:FacialDic}  \textbf{(B)}.}  
    \label{fig:RT}
\end{figure*} 

The real-time and on-the-edge flow diagram is presented in \cref{fig:RT}. 
The flow diagram is split into two specific applications; Eating and Facial Muscle Movement recognition. 

\textit{Eating Scenario:} TensorFlow Lite for MCU was used to generate the embedded version of the NN models. 
For RTE recognition, two algorithms were used. 
In \cref{fig:RT}\textbf{A} is the eating/drinking flow diagram. 
A sliding window of 2 seconds (100 samples) with a step size of 0.5s is used as an input data frame to the NNs.
For the eating monitoring, the PC is 0.5489 Watts (only MMG-model), and when the inertial model is activated, the PC is 0.5988 Watts.

\textit{Facial Scenario:} The \cref{fig:RT}\textbf{B} depicts the procedure for the facial expressions' case. 
The first step consists of movement detection (using acceleration), reducing power consumption by 16\% (from 0.55 to 0.46 Watts).
The movement detection is based on a threshold condition ruled by $\Sigma_{n=0}^{5}=|{a_x}|_n + |{a_y}|_n +|{a_z}|_n$. 
The motion detection only applies to the facial scenario as depicted in \cref{fig:RT}\textbf{B}. 
Then, the data collection will run until no movement is detected. 
Therefore, the size of the input window is variable and depends on the duration of the detected movements. 
The NN input is fixed at 100-time samples, so dynamic resampling of the window size is necessary. 
After data collection, the data is resampled to 100 samples using the equation: $Y_i = (p*a_{index+1} + (NS-p)*a_{index})/NS-1$. Where $NS=$new sampling, $OS=$old sampling, $p=i*OS\%NS$, $index=i*(OS/NS)$ for $i\in{(0,NS-1)}$.
The resample's output is the input to the MMG model.
In the case of $activity \neq Null$, the inertial model will output the recognized facial expression. 
The power consumption (PC) for the facial expression solution is 0.55 Watts and 0.65 Watts when MMG and inertial model are activated.
\footnote{We used the USB Digital Power Meter: available in  \url{https://www.az-delivery.de/en/products/charger-doktor} DLA: April 03, 2024.}

%% file: Chapters/ResultsDiscussion.tex
\section{Results and Discussion}
\label{sec:Results}
\begin{figure*}
    \centering  
    \includegraphics[width=\textwidth]{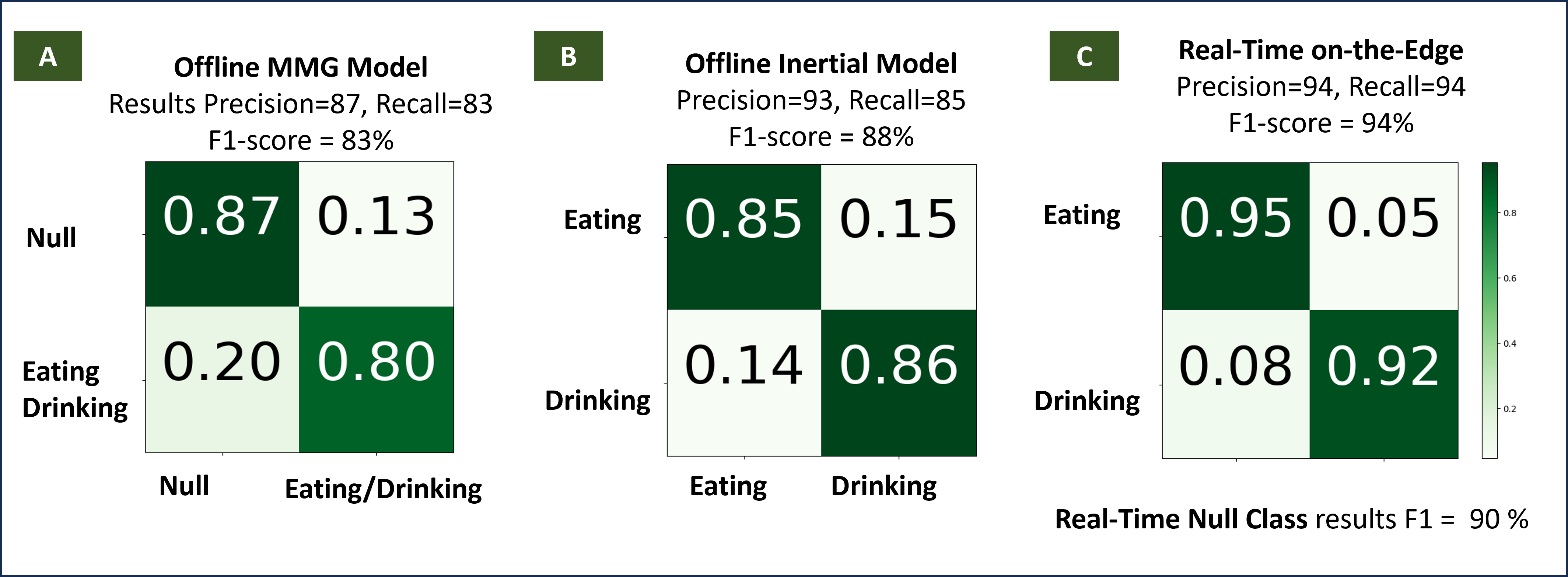}
    \caption{Results of the offline MMG-Model with Five Volunteers (Leave-one-session-out cross-validation) in Lunch/Dinner Scenario;F1-score=83 \%\textbf{(A)}. 
    Results of the offline Inertial-Model with Five Volunteers (Leave-one-session-out cross-validation) Lunch/Dinner Scenario; F1-score=88 \%\textbf{(B)}. 
    Real-Time on-the-Edge Recognition Results for Five Unseen Volunteers (User-independent) in Snacking Scenario; F1-score = 94 \%\textbf{(C)}. 
    }
    \label{fig:ResultEat}
\end{figure*}
\begin{figure*}
    \centering  
    \includegraphics[width=\textwidth]{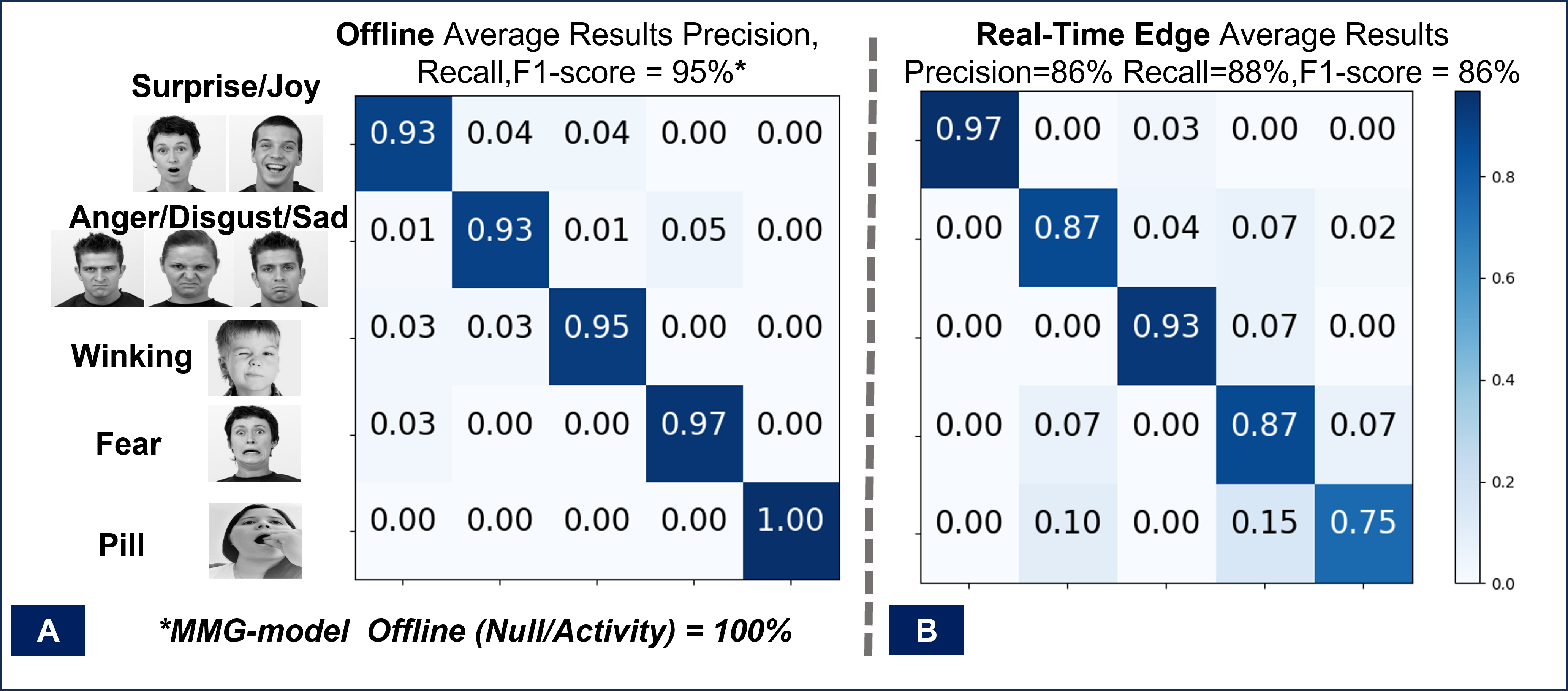}
    \caption{Results of the offline Inertial-Model for Ten Sessions on Different Days with Leave-One-Session Out Cross Validation for the Recognition of the Dictionary in \cref{fig:FacialDic}; Joy/Surprise(1), Anger/Disgust/Anger(2), Winking(3), Fear(4) and taking a pill(5) and F1-score=95\%\textbf{(A)}. 
    Real-Time and on-the-Edge Results of the Inertial-Model for Three Sessions on Different Days for the Recognition of the Facial Activities in the Dictionary in \cref{fig:FacialDic}; F1-score=86\%\textbf{(B)}.
    }
    \label{fig:ResultFacial}
\end{figure*}
The test group and the training group have the same size and sex distribution but with no overlap between participants. 
\cref{fig:ResultEat}\textbf{A-B} shows the eating/drinking (5 persons, 4-fold cross-validation) offline results for the collaborative approach; MMG model (Null vs Activity) \cref{fig:ResultEat}\textbf{A} and Inertial model (Eating vs Drinking) \cref{fig:ResultEat}\textbf{B}. 
The results in \cref{fig:ResultEat}\textbf{A-B} are from the training group (three females and two males) in a leave-one-session out cross-validation scheme. 
In \cref{fig:ResultEat}\textbf{C} are the results for the real-time and on-the-edge evaluation, performed by the test group. 
For the RTE, the test group (three women and two men) ate and drank freely without any restrictions or instructions.  
The RTE for the eating/drinking monitoring F1-score was 94\% with five additional subjects (test group), not within the training group.  
A total of ten participants performed the eating/drinking scenario evaluation. 
Hence, our approach could generalize the data from different participants, but more unknown participants are still required.

In \cref{fig:ResultEat}\textbf{B}, the offline results of the classification between eating and drinking instances with a score F1 = 88\% are displayed. 
Instances are defined with a window size of 2 seconds and a step size of 0.5 seconds. 
\cref{fig:ResultEat}\textbf{C} shows the online results of the NN network embedded in the glasses with a score F1 = 94\%.
In the online evaluation, a majority voting system was used to classify the eating/drinking episodes.
The voting buffer consists of five windows of size 2 seconds and step size 0.5 seconds, for a total inference time of about 4 seconds. 
Therefore, we can assume that the voting mechanism is the main reason for the improved performance of 6\%.

A 10-fold cross-validation with a leave-one session out (10 sessions on different days) was performed to obtain the offline results in \cref{fig:ResultFacial}\textbf{A}.  
The real-time and on-the-edge results for the facial expression scenario (3 sessions, different days, one volunteer) are shown in \cref{fig:ResultFacial}\textbf{B}. 
There was a reduction of 10\% in the F1-score between the offline and the embedded solution.
We believe this is due to errors in the motion detection algorithm in conjunction with the simplified (linear-based) dynamic resampling technique deployed in the prototype compared to the Fourier-based dynamic resampling of the training set.

The class of taking a pill degrades the most in performance between online and offline results, from 100\% to 75\%.
In the offline results, the start and end of facial movements are manually annotated using the videotaped sessions as ground truth. 
In contrast, in online recognition, the start and end of facial movements depend on the motion detection algorithm. 
The motion detection algorithm is based on a threshold to determine the transition from a static to a moving state. 
Most classes are composed of a strong movement to make the facial expression, then a static period, and return to a neutral face with another strong movement. 
These steps are easily detected by the motion detection algorithm. 
But, the gesture of taking a pill is more complex in comparison. 
The activity of taking a pill involves a strong movement of the hand toward the face, followed by a slow movement of inserting the pill into the mouth, and ending with a strong movement of the hand coming back to rest. 
Thus, the gesture of taking a pill has a semi-static state (inserting the pill) compared to the other categories with a more defined static state. 
Therefore, we believe that the motion detection algorithm is the main cause of a reduction in the performance of 25\% in the gesture of taking a pill. 
A future solution could be to deploy two different motion detection algorithms and make a probabilistic voting decision for classification.

The results (both scenarios) have an F1-score $\geq$ 86 \%, meaning that our approach holds promise for further development. 
This work has demonstrated the feasibility of using the MeciFace idea in two scenarios. 
Future work can focus on merging the two scenarios into one to obtain contextual information about users during sporadic eating activities.
Besides, it would be meaningful to exploit the additional sensing information with the barometer/gas sensor and the microphone. 
For example, the gas sensor can detect volatile organic compounds (VOCs), volatile sulfur compounds (VSCs), and carbon monoxide, among other gases. 
Detecting VSCs is an indicator of bacteria growing. 
In \cite{OralCancerVSC}, the authors use the VSCs in exhaled breath as a potential diagnostic method for oral cancer. 
On the other hand, the audio information can be used to detect sound related to emotions as a contextual source \cite{bello2020facial}. 

Integrating IMU sensors into smart glasses is not a challenge as they are already unobtrusively integrated into a commercial smart wearable like in \cite{OCOsense}. 
The FSR and Piezo sensors are also straightforward to integrate due to their flexible design and minimal circuit requirement with only an analog to digital input constraint per sensor. 
Additionally, the mechanomyography sensing modality is completely passive compared to the IMU-based modality, so the power consumption is kept low.

\textbf{Limitations}
Here we present a list of the limitations we have identified as well as future directions for optimizing the system:

\textit{Experiment Extension}. The evaluation is considered preliminary with only one user for the facial and ten volunteers for the eating. 
Thus, an extended experiment setting is crucial to demonstrate the generality of the approach. 
The experiment should include variability in sex and culture to have fair training of the NNs to recognize facial activities as in \cite{bello2023inmyface}. 

\textit{Neural Network Tuning}. The NNs were deployed using the TensorFlow Lite framework for MCU without any additional optimization technique. 
It is relevant to explore optimization approaches such as; quantization aware training, pruning, and quantization in the bits level, to improve the power consumption and the performance of the NNs.
These optimization techniques are highly dependent on the selected embedded hardware. 
We leave this exploration for future work.

\textit{Miniaturization}. The hardware in this work is a fast prototype, but the selected sensors have reduced dimensions, as shown in \cref{table:Sensors}, which could be fully embedded in the glasses frame to improve comfort. 

\textit{Human Feedback}. After all the above limitations are addressed, it is crucial to do a human study to expose the weaknesses of the design and tune it to include user perception.  

%% file: Chapters/Conclusion.tex
\section{Conclusion}
\label{sec:Conclusion}
In this paper, we proposed MeciFace, an innovative energy-efficient wearable system for real-time facial and eating activity recognition. 
By leveraging a glass frame as the wearable accessory, we strategically deployed sensors and the microcontroller, ensuring minimal intrusion into the user's daily life. 
The fusion of mechanomyography and inertial information on eyeglass temples and nose bridges allowed for comprehensive monitoring of facial expressions and eating activities.
In the experimental results, we demonstrated the performance of MeciFace in real-time and on-the-edge, achieving an F1 score of $\geq86\%$ for the facial expressions scenario and an F1 score of 94\% for the eating scenario with a test group of five volunteers (user-independent case). 
Two groups of volunteers were recruited for the evaluation.
The training group and the test group have the same size and sex composition, but the participants do not overlap.
For the training group, five volunteers (three female and two male) participated in the eating monitoring experiment in a natural setting during lunch/dinner.
The second group of five volunteers (three women and two men) were recruited for the real-time and on-the-edge evaluation of the eating monitoring case. 
Hence, with this methodology, our results are user-independent and sex-balanced, with high variability across cultures.

The hierarchical scheme implemented in the system significantly reduced power consumption, maintaining it below 0.55 Watts, thus enhancing the wearability and practicality of the device. 

The TensorFlow Lite for Microcontroller framework enabled the seamless deployment of neural network-based models in their embedded versions.
Techniques such as quantization and pruning can contribute further to memory reduction and efficient utilization of embedded resources, ensuring the system's sustainability.
MeciFace can be easily extended to include contextual information from the environment, thanks to the incorporation of barometer/gas sensors and a microphone on the glasses' nose bridge.
This extension enhances the potential of the system to detect stress-triggered eating episodes and offers a holistic approach to monitoring emotional eating behaviors.